\documentclass[lettersize,journal]{IEEEtran}
\usepackage{amsmath,amsfonts}
\usepackage{algorithmic}
\usepackage{algorithm}
\usepackage{array}
\usepackage[caption=false,font=normalsize,labelfont=sf,textfont=sf]{subfig}
\usepackage{textcomp}
\usepackage{stfloats}
\usepackage{url}
\usepackage{verbatim}
\usepackage{graphicx}
\usepackage{cite}
\usepackage{multirow}
\usepackage{pifont}
\usepackage{makecell}
\begin{document}

\title{CLIPVehicle: A Unified Framework for Vision-based Vehicle Search}

\author{Likai Wang, Ruize Han, Xiangqun Zhang, and Wei Feng
\thanks{
This work was supported by the National Natural Science Foundation of China (NSFC) under Grant 62072334, 62402490 and the Shenzhen Basic Research Foundation under Grant JCYJ20240813154920027. \textit{(Corresponding author: Wei Feng. E-mail: wfeng@ieee.org.)}

Likai Wang, Xiangqun Zhang, and Wei Feng are with the College of Intelligence and Computing, Tianjin University, Tianjin 300350, China.

Ruize Han is with the Shenzhen Institute of Advanced Technology, Chinese Academy of Sciences, Shenzhen, China, and the Shenzhen University of Advanced Technology. 
} 
}



\maketitle

\begin{abstract}
Vehicles, as one of the most common and significant objects in the real world, the researches on which using computer vision technologies have made remarkable progress, such as vehicle detection, vehicle re-identification, \textit{etc}.
To search an interested vehicle from the surveillance videos, existing methods first pre-detect and store all vehicle patches, and then apply vehicle re-identification models, which is resource-intensive and not very practical.
In this work, we aim to achieve the joint detection and re-identification for vehicle search.
However, the conflicting objectives between detection that focuses on shared vehicle commonness and re-identification that focuses on individual vehicle uniqueness make it challenging for a model to learn in an end-to-end system.
For this problem, we propose a new unified framework, namely CLIPVehicle, which contains a dual-granularity semantic-region alignment module to leverage the VLMs (Vision-Language Models) for vehicle discrimination modeling, and a multi-level vehicle identification learning strategy to learn the identity representation from global, instance and feature levels.
We also construct a new benchmark, including a real-world dataset CityFlowVS, and two synthetic datasets SynVS-Day and SynVS-All, for vehicle search.
Extensive experimental results demonstrate that our method outperforms the state-of-the-art methods of both vehicle Re-ID and person search tasks.
We will release the benchmark and code proposed in this work to the public.
\end{abstract}


\section{Introduction}
\label{sec:intro}
Vehicle is one of the most significant category of objects in the real world. The research on vehicles has many practical applications, such as intelligent transportation systems (ITS), automatic driving, \textit{etc.}
To build the intelligent transportation systems, computer vision based system plays an important role.
In the area of computer vision, many vehicle-related topics have been studied, including the vehicle detection~\cite{deshmukh2024multi,wang2022review}, vehicle counting~\cite{guo2023scale,gao2024nwpu}, vehicle 3D tracking~\cite{li2022bevformer,zhang2022mutr3d}, and vehicle re-identification (Re-ID)~\cite{zhang2024multimodality,amiri2024comprehensive,li2024day}.

\begin{figure}[t]
	\centering
	\includegraphics[width=1.0\linewidth]{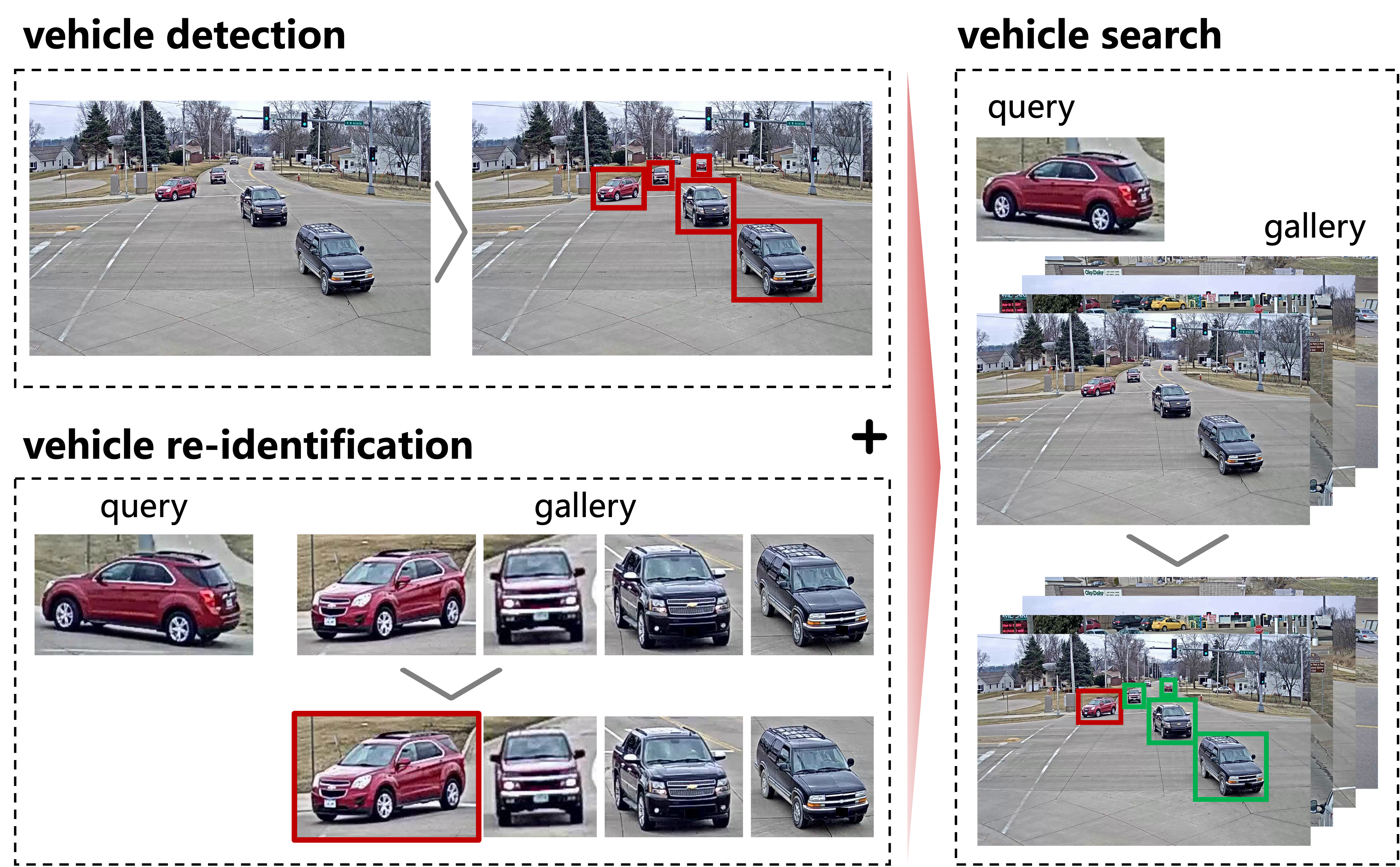}
	\caption{Illustration of vehicle detection, vehicle re-identification (Re-ID), and the proposed end-to-end vehicle search tasks. 
Vehicle detection (upper left) aims to detect all vehicles (red rectangles) in the input frame without considering the vehicle identity information.
Vehicle Re-ID (lower left) aims to identify the same vehicle as query from gallery images (red rectangle), requiring the pre-cropped vehicle patches as input.
The proposed end-to-end vehicle search (right) jointly tackle the vehicle detection and Re-ID, and aims to retrieve the same vehicle as query (red rectangle) from all detected boxes (red and green rectangles), which is more practical and challenging.}
	\label{fig:vs}
\end{figure}

In practice, a typical application is to search a target vehicle (based on a given vehicle image) from surveillance videos in traffic scene. This task has great practical significance, but has been overlooked and still lacks a unified solution framework.
Specifically, to achieve this, existing methods~\cite{liu2017beyond,liu2019pvss} first use detection networks to obtain all vehicle patches appearing in the videos, which then serve as input of vehicle re-identification to build the search gallery.
This is not very practical since it requires to detect and save all vehicle boxes in the whole video in advance.
Also, the two separate networks for handling detection and re-identification would lead to heavy network architecture and prevent them from joint optimization.
To be more practical, in this work, we aim to achieve joint detection and re-identification for vehicle search.
As shown in Fig.~\ref{fig:vs}, given a target vehicle (as query) and raw video frames (as gallery), vehicle search should retrieve all vehicles (in terms of bounding box) with the same ID as query.

We find that, different from the vehicles, person search has been studied for a long time with abundant literatures~\cite{wang2022person,lin2021person}. This demonstrates that \textit{the search problem is worthy of study but has been neglected for vehicles}.
However, compared to person search, vehicle search has met some new difficulties.
Specifically, the identification feature construction manners for vehicle are not very rich like persons, such as the human pose~\cite{rao2023transg}, human gait~\cite{wang2022benchmark}, \textit{etc.}
Also, the vehicle appearance is not as discriminative as the humans, which have various outfits and body shapes.
This way, how to learn the discriminative features to identify the same/different vehicles for vehicle search, is more difficult.

We summary the main challenges to build a unified framework for vehicle search. 
1) In vehicle search, the vehicle detection and re-identification sub-tasks have fundamentally different objectives, \textit{i.e.}, the detection task aims to distinguish vehicles from the background, focusing on \textit{shared vehicle commonness}, while the re-identification task aims to distinguish different vehicle IDs, focusing on \textit{individual vehicle uniqueness}. 
2) Detection, as the prerequisite task, its results directly effect the learning efficiency and capability of subsequent re-identification model, \textit{i.e.}, errors in detection are propagated and can severely affect the performance of the re-identification.

To address the above challenges, we propose a new end-to-end unified framework namely CLIPVehicle for vehicle search.
Specifically, to reconcile the vehicle detection and re-identification tasks with different objectives, we consider to integrate the multi-modal pre-trained models, \textit{e.g.}, the CLIP~\cite{radford2021learning}, CoOp~\cite{zhou2022learning}, for obtaining abundant semantic descriptions.  
We propose a dual-granularity semantic-region alignment module.
This module makes use of the vision-language cross-modal alignment ability to improve the vehicle search, where we consider both the object-granularity and ID-granularity text prompts from different granularities, to handle the  vehicle detection and re-identification tasks, respectively.
Also, to achieve effective and efficient training of the re-identification model less affected by the inaccurate detection results, besides commonly used re-identification head, we further design a multi-level vehicle identification learning strategy. This effectively enhances the vehicle identification ability from the global image (multi-instance) level, the ground-truth bounding box (single-instance) level and the feature level.
The main contributions of this work are:

\begin{itemize}
	\item This is the pioneer work to study the vision-based vehicle search problem in a unified framework, which jointly tackles the vehicle detection and re-identification. This problem is more practical and promotes the study of vehicle identification a further step.
	\item We propose a new end-to-end unified framework namely CLIPVehicle for the vehicle search problem, which contains a dual-granularity semantic-region alignment module to leverage the VLMs for vehicle discrimination modeling, and a multi-level identification learning strategy to learn the vehicle ID representation from different levels.
	\item We build a new benchmark for vehicle search based on several existing vehicle datasets. Extensive experimental results demonstrate that our method outperforms the state-of-the-art methods for both vehicle Re-ID and person search tasks.
\end{itemize}

\section{Related Work}
\label{sec:related}

\textbf{Vehicle detection and Re-ID.}
Due to the essential role of vehicles in intelligent transportation systems (ITS), vehicle-related researches, such as vehicle detection~\cite{mittal2023ensemblenet,wang2022review} and vehicle re-identification~\cite{khorramshahi2023robust,amiri2024comprehensive,li2024day}, have attracted widespread attention.
Vehicle detection aims to locate vehicles in raw images, serving as a basis for ITS.
Existing vehicle detection methods based on deep learning can be divided into two categories, \textit{i.e.}, two-stage and one-stage~\cite{wang2022review}.
Two-stage methods~\cite{chan2023influence,chaudhuri2024smart}, developed based on Faster R-CNN~\cite{ren2016faster}, \textit{etc.}, first generate region proposals and then find vehicle targets from the region proposals.
One-stage methods~\cite{kang2024yolo,pan2024lvd}, developed based on YOLO series~\cite{redmon2016you} and SSD~\cite{liu2016ssd}, directly regress the position of vehicles from the original images.

{Vehicle Re-ID} aims to identify the same vehicle across non-overlapping cameras.
Several datasets, including VeRi-776~\cite{liu2016large}, VERI-Wild~\cite{lou2019veri}, Vehicle-1M~\cite{guo2018learning}, \textit{etc.}, have been proposed to support this task.
Based on them, most existing methods focus on either discriminative feature learning~\cite{chen2023global,sun2024heterogeneous} or deep metric learning~\cite{ghosh2023relation,qiu2024camera}.
The former adopts CNNs or transformers to learn global features from the whole vehicle images~\cite{zhu2018shortly,zheng2020vehiclenet}, or learn local features using keypoint locations~\cite{hu2023vehicle} and image partitions~\cite{shen2023git,sun2024heterogeneous}.
The latter utilizes contrastive loss~\cite{yu2023weakly,qiu2024camera} or triplet loss~\cite{ghosh2023relation,zhao2024benchmark} to map vehicle images into a feature space, where vehicles with the same ID are as close to each other as possible, while vehicles with different IDs are farther away from each other.
However, the above methods start from pre-cropped vehicle images, with manually annotated or automatically detected.
Despite the remarkable progress in both vehicle detection and vehicle Re-ID, the two tasks are studied independently of each other all the time, and have not been considered in a unified framework for application. In this work, we aim to bridge the detection and Re-ID tasks for vehicle search, which is more practical for real-world applications.

\textbf{Vehicle search} aims to locate and identify a target vehicle from raw surveillance videos, which has been rarely studied.
Existing works~\cite{feris2011attribute,liu2019pvss} first detected and tracked vehicles over the video, and then extracted fine-grained attributes and ingested them into a database to allow future search, where \textit{the detection and Re-ID stages are performed separately}.
In addition, natural language-based vehicle search has been preliminarily explored in Lee \textit{et al.}~\cite{lee2021sbnet}, which proposed a segmentation-based model SBNet to find target vehicles within a given target image using a natural language description as a query.
However, they \textit{retrieve all vehicles matching the description in surveillance videos rather than unique target vehicle}.
Also, the natural language descriptions not only require additional annotations, but also contain less information than images.

\textbf{Person search.}
Pedestrians, as the most common objects in computer vision, have received more attention than vehicles~\cite{lin2021person,zhang2024towards}. Xu \textit{et al.}~\cite{xu2014person} first introduced the person search by jointly modeling the commonness of people and the uniqueness of a specific person. 
In recent years, large-scale person search datasets, \textit{e.g.}, CUHK-SYSU~\cite{xiao2017joint} and PRW~\cite{zheng2017person}, have been proposed successively, promoting the rapid development of this task.
Existing person search methods can be divided into three categories, \textit{i.e.}, two-step, parallel end-to-end, and sequential end-to-end.
Two-step methods~\cite{wang2022person,wang2020tcts} employ two separate models for person detection and Re-ID, which are time-consuming and resource-consuming.
Parallel end-to-end methods~\cite{chen2021norm,dong2020bi} first employ the region proposal network to generate region proposals, and then feed them into parallel detection and Re-ID branches.
The main challenge faced by such methods is that the input of ID head is the low-quality proposals, which affects the learning of fine-grained identity-related features.
Sequential end-to-end methods~\cite{li2021sequential,zhang2024joint} share the stem representations of person detection and Re-ID, and tackle the two subtasks sequentially with two separate heads, achieving a trade-off between accuracy and efficiency.
In this work, we follow the sequential end-to-end (ReID-by-detection) pipeline to learn the semantic-guided and coarse-to-fine features for joint vehicle detection and Re-ID, \textit{i.e.}, vehicle search.

\textbf{Vision-language model learning.}
Recently, the emergence of large vision-language models (VLMs), such as CLIP~\cite{radford2021learning}, has opened new opportunities for computer vision tasks.
With 400 million text-image pairs for initial pre-training, CLIP exhibits a superior ability in many downstream tasks, \textit{e.g.}, image classification~\cite{abdelfattah2023cdul}, object tracking~\cite{li2023ovtrack}, semantic segmentation~\cite{zhou2023zegclip}.
Further, to address the challenge of hand-crafted prompt templates, inspired by the advances in prompt learning research in NLP (natural language processing), Zhou \textit{et al.}~\cite{zhou2022learning} proposed Context Optimization (CoOp) to make text prompts adaptively learnable.
Recent methods CLIP-ReID~\cite{li2023clip} and CCLNet~\cite{chen2023unveiling} have integrated CoOp into the image Re-ID task, and verified that it can effectively help capture fine-grained ID-related features.
Inspired by that, in this work, we propose to equip pre-trained VLMs, \textit{e.g.}, CLIP, CoOp, into the proposed vehicle search framework, aiming to adaptively and effectively learn the attribute-based vehicle features (\textit{e.g.}, vehicle color, type).

\section{Proposed Method}

\begin{figure*}[t]
\centering
\includegraphics[width=1.\linewidth]{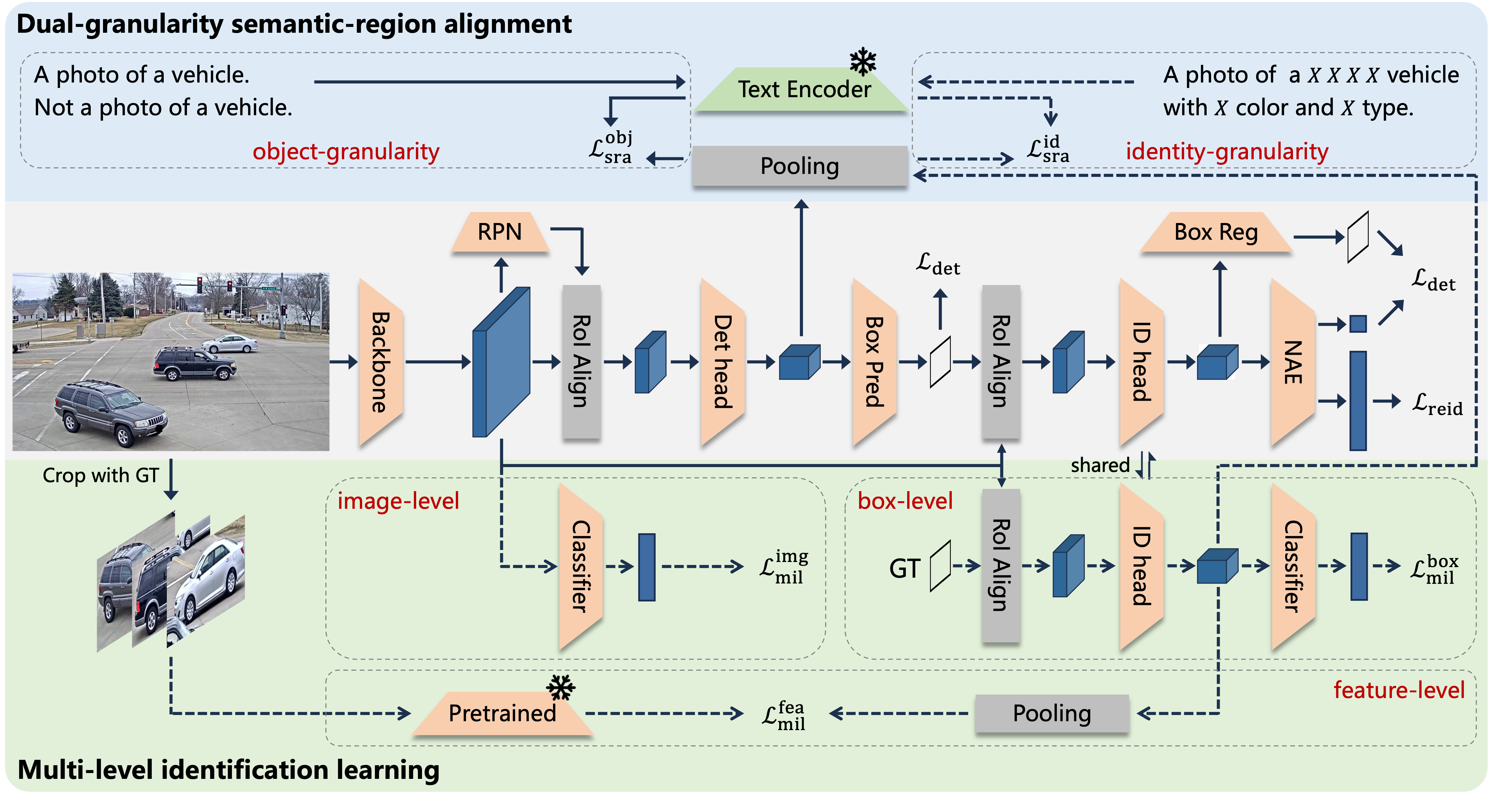}
\caption{Overview of the proposed end-to-end unified framework CLIPVehicle for vehicle search (training stage). Given a raw frame, we sequentially handle the vehicle detection and re-identification sub-tasks (middle part). For more discriminative features learning, dual-granularity semantic-region alignment is proposed to leverage object-granularity and identity-granularity text prompts to guide detection and re-identification, respectively (upper part). Also, multi-level identification learning strategy is proposed to further enhance the model capability, comprising image-level, box-level, and feature-level in a coarse-to-fine manner (lower part).}
\label{fig:overview}
\end{figure*}

\subsection{Problem formulation}
In this work, we focus on a new task namely end-to-end vehicle search. We first provide the problem formulation of it.
Formally, in the training stage, the vehicle search training set can be defined as $\mathcal{D} = \{(\mathbf{x}_t,\mathbf{y}_t) \}_{t=1}^{T}$, where $\mathbf{x}_t$ denotes the $t$-th raw frame, and $\mathbf{y}_t$ denotes the corresponding annotation, $T$ is the total number of frames in $\mathcal{D}$.
Specifically, $\mathbf{y}_t=\{(b_t^j,c_t^j)\}_{j=1}^{N^{t}}$, 
where $N^{t}$ denotes the number of vehicles within frame $\mathbf{x}_t$, $b_t^j$ denotes the ground-truth bounding box and $c_t^j\in \{1,2,\cdots, C\} $ is the identity class label of $j$-th vehicle in $\mathbf{x}_t$. 
$C$ denotes the number of vehicle identities in training set $\mathcal{D}$.
Given $\mathcal{D}$, vehicle search aims to train a model to detect the boxes of vehicles in raw frames and extract their discriminative identity-related features, the similarity between which of boxes belonging to the same vehicle is expected to be maximized.

During evaluation, given a cropped query vehicle image $\mathbf{q}$ and a gallery set $\mathcal{G}=\{\mathbf{x}_k\}_{k=1}^{K}$, where $K$ is the number of frames in $\mathcal{G}$, the learned model is required to return a series of detected vehicle box candidates with a descending ranking list according to the identification consistency with $\mathbf{q}$.
This can be achieved by calculating the similarity between features of all detected boxes in $\mathcal{G}$ and feature of the query $\mathbf{q}$.
As the vehicle search results, the detected boxes, which have an IoU larger than 0.5 with ground-truth boxes sharing the same identity label with the query, should be ranked higher.
Note that the training set and test set do not have overlapping vehicle identities, like vehicle Re-ID problem.

\subsection{Overview}
For the vehicle search problem, we propose an end-to-end unified framework CLIPVehicle.
As shown in the middle part of Fig.~\ref{fig:overview}, given a raw frame $\mathbf{x}_t$ as input, we first extract the visual feature map as $\mathbf{F}_t=f_{\mathrm{b}}(\mathbf{x}_t|\theta )$, which is shared by both detection and re-identification head. $f_{\mathrm{b}}(\cdot |\theta )$ denotes the backbone network with parameters $\theta$.
We then apply the RPN~\cite{ren2016faster} to generate a set of potential region proposals $\mathcal{P}_t \subset \mathbb{R}^{4 \times p} $, $p$ is the number of proposals.
For each proposal $i$, we perform the RoI Align operation on $\mathbf{F}_t$ to extract proposal feature map $\mathbf{f}_i^{\mathrm{pro}} \in \mathbb{R}^{h\times w \times d}$, which $h$, $w$, $d$ denote the height, width, the number of channels of the feature maps, respectively.
After that, the detection head $f_{\mathrm{det}}(\cdot)$ and box predictor $f_{\mathrm{box}}(\cdot)$~\cite{ren2016faster} are applied to make box predictions $\mathcal{B}_i$ from per-region features
\begin{equation}
  \mathbf{f}_i^{\mathrm{det}} = f_{\mathrm{det}}(\mathbf{f}_i^{\mathrm{pro}}), \quad \mathcal{B}_i=f_{\mathrm{box}}(\mathbf{f}_i^{\mathrm{det}}),
  \label{eq:boxpred}
\end{equation}
where $\mathbf{f}_i^{\mathrm{det}} \in \mathbb{R}^{\frac{h}{2}\times \frac{w}{2} \times 2d}$, and $\mathcal{B}_i \in \mathbb{R}^5$ denotes the (preliminary) detection prediction results, including the predicted box and the foreground-background classification (be a vehicle or not) score.
The widely-used detection loss $\mathcal{L}_\mathrm{det}$, including a bounding box regression and foreground-background classification, is adopted on $\mathcal{B}_i$ for supervision.

Based on the predicted boxes $\mathcal{B}_i$, we first perform the RoI Align operation on $\mathbf{F}_t$ to extract box feature maps $\mathbf{f}_i^{\mathrm{box}} \in \mathbb{R}^{h\times w \times d}$, which is next mapped into $\mathbf{f}_i^{\mathrm{id}} \in \mathbb{R}^{\frac{h}{2}\times \frac{w}{2} \times 2d}$ by identification (ID) head $f_{\mathrm{id}}(\cdot)$.
We then apply the bounding box regression head $f_{\mathrm{reg}}(\cdot)$ and Norm-Aware Embedding (NAE) $f_{\mathrm{nae}}(\cdot)$~\cite{chen2021norm} to refine box predictions and extract discriminative identity-related features as
\begin{equation}
  {\mathcal{B}}_i^{\mathrm{box}} = f_{\mathrm{reg}}(\mathbf{f}_i^{\mathrm{id}}), \quad (\mathbf{f}_i^{\mathrm{nae}}, \mathcal{B}_i^{\mathrm{cls}})=f_{\mathrm{nae}}(\mathbf{f}_i^{\mathrm{id}}),
  \label{eq:reid}
\end{equation}
where ${\mathcal{B}}_i^{\mathrm{box}} \in \mathbb{R}^4$ is the refined box result, $\mathcal{B}_i^{\mathrm{cls}} \in [0,1]$ is to estimate the box is a vehicle or not.
Similar with ${\mathcal{B}}_i$, the detection loss $\mathcal{L}_\mathrm{det}$ is applied to ${\mathcal{B}}_i^{\mathrm{box}}$ and $\mathcal{B}_i^{\mathrm{cls}}$, for supervision.
Finally, the output ID feature after NAE, \textit{i.e.}, $\mathbf{f}_i^{\mathrm{nae}} \in \mathbb{R}^o$, is used to ID classification.
We apply the Online Instance Matching (OIM) loss function~\cite{xiao2017joint} on $\mathbf{f}_i^{\mathrm{nae}}$ as the ID discrimination loss $\mathcal{L}_\mathrm{reid}$.

During the training phase, inspired by the superior ability of VLMs in many computer vision tasks, we introduce the pre-trained CLIP~\cite{radford2021learning} and prompt learning method CoOp~\cite{zhou2022learning} into vehicle search, and propose a dual-granularity semantic-region alignment strategy, using both the \textit{object-granularity and identity-granularity language prompts} to handle the vehicle detection and re-identification tasks, respectively (will be introduced in Sec.~\ref{sec:dual}).
Besides, in order to guide the model to learn more discriminative vehicle identification features, we propose a multi-level identification learning strategy that focuses on the \textit{image-level, box-level, and feature-level information from coarse to fine} (will be introduced in Sec.~\ref{sec:multi}).

\subsection{Dual-granularity semantic-region alignment}
\label{sec:dual}
In this section, we propose to integrate the pre-trained vision-language model CLIP~\cite{radford2021learning} to vehicle search, transferring the additional knowledge extracted from semantic descriptions to vehicle search models via feature alignment.
CLIP consists of a text encoder that extracts text embedding from natural language and a image encoder that extracts visual embedding from high-dimensional images. It has achieved the cross-modal alignment of the two embedding spaces through utilizing 400 million image-text pairs for training.
When adapting CLIP for the downstream task, \textit{i.e.}, vehicle search in this work, we aim to achieve the alignment between image embeddings of regions and text embeddings of descriptions that the regions relate to.
Since vehicle search jointly handles two tasks with different objectives, \textit{i.e.}, vehicle detection that focuses on the distinction between vehicle objects and background, and vehicle re-identification that focuses on the distinction between each vehicle and other vehicles, we propose the dual-granularity semantic-region alignment, designing object-granularity and ID-granularity text prompt to match the two objectives, respectively (shown in upper part of Fig.~\ref{fig:overview}).

\textbf{Foreground-background discrimination.}
We first propose the object-granularity text prompt for the distinction between foreground (vehicle objects) and background.
Specifically, we use a simple prompt strategy, \textit{i.e.}, the text prompt of foreground is defined as `A photo of a vehicle', and the text prompt of background is defined as `Not a photo of a vehicle'.
The two prompts are fed into the frozen text encoder of CLIP to obtain the text embeddings $\mathbf{t}^\mathrm{fore}$ and $\mathbf{t}^\mathrm{back}$.
For each pooled region feature $\mathbf{f}_i^{\mathrm{det}}$ output by detection head, we calculate the object-granularity semantic-region alignment loss as
\begin{equation}
	\sigma _1 = \exp (\mathrm{sim}(\mathbf{f}_i^{\mathrm{det}}, \mathbf{t}^\mathrm{fore})),  \sigma_2=\exp (\mathrm{sim}(\mathbf{f}_i^{\mathrm{det}}, \mathbf{t}^\mathrm{back})),  \label{eq:ali1-1}
\end{equation}
\begin{equation}
  \mathcal{L}_\mathrm{sra}^\mathrm{obj}=\sum _i -\log \frac{c_{i}^\mathrm{det}\sigma_1 + (1-c_{i}^\mathrm{det})\sigma_2}{\sigma_1 + \sigma_2 },
  \label{eq:ali1-2}
\end{equation}
where $\mathrm{sim}(\cdot, \cdot)$ measures the similarity between two feature vectors, the value of $c_{i}^\mathrm{det}$ is 1 if the corresponding region proposal matches a ground-truth bounding box (with IoU larger a threshold, considered as vehicle object), else 0 (considered as background).

\textbf{Vehicle identity classification.}
We then propose the ID-granularity text prompt for vehicle ID classification.
Different from object classification, where labels are described as natural language, the labels for ID classification are integers and do not have actual semantic information.
To solve this problem, we introduce the prompt learning method CoOp~\cite{zhou2022learning} to learn ID-specific tokens for text descriptions.
Specifically, the text prompt is defined as `A photo of a $[X]_1[X]_2\cdots [X]_M$ vehicle', where $[X]_m$ is the \textit{learnable text token} with random initialized parameters.
Moreover, compared to other objects like pedestrians, vehicles have smaller inter-class difference (different vehicle looks similarly) and larger intra-class difference (due to the frequent view-point difference).
For more discriminative vehicle feature extraction, recent vehicle Re-ID works~\cite{li2022attribute,zhang2024aivr} attempt to exploit vehicle attribute cues, \textit{i.e.}, colors and types (sedan, van, suv, truck, \textit{etc.}), which have been proven to be effective.
Inspired by that, we expand the text prompt to `A photo of a $[X]_1[X]_2\cdots [X]_M$ vehicle with $[X]$ color and $[X]$ type', to explore attribute-based information helpful for vehicle ID classification.
Note that, the tokens are learned in an additional pre-training stage, and are frozen when training vehicle search model.

The learned ID-specific text descriptions are fed into the frozen text encoder of CLIP to obtain the text embeddings $\mathbf{t}_c$ for each ID, where $c \in \{1,2,\cdots , C\} $ denotes the $c$-th identity class.
For each pooled region feature $\mathbf{f}_j^{\mathrm{gt}}$ output by ID head (features of ground-truth boxes, described in Sec.~\ref{sec:multi}), we calculate ID-granularity alignment loss as
\begin{equation}
  \mathcal{L}_\mathrm{sra}^\mathrm{id}=\sum _j \sum _{c=1} ^ C -\alpha \log \frac{\exp (\mathrm{sim}(\mathbf{f}_{j}^\mathrm{gt}, \mathbf{t}_c))}{\sum _{q=1} ^ C \exp (\mathrm{sim}(\mathbf{f}_{j}^\mathrm{gt}, \mathbf{t}_q))},
  \label{eq:ali2}
\end{equation}
where $\alpha$ is set as 1 if the index $c$ equals to the corresponding identity class of region feature $\mathbf{f}_j^{\mathrm{gt}}$, else 0.

In total, the optimization objective of dual-granularity semantic-region alignment is defined as $\mathcal{L}_\mathrm{sra} = \mathcal{L}_\mathrm{sra}^\mathrm{obj}+\mathcal{L}_\mathrm{sra}^\mathrm{id}$.

\subsection{Multi-level identification learning}
\label{sec:multi}
To further enhance the capability of vehicle search model, we propose a multi-level identification learning strategy comprising image-level, box-level, and feature-level, in a coarse-to-fine manner (shown in lower part of Fig.~\ref{fig:overview}).

\textbf{Image-level (global) multi-ID classification loss.}
Different from vehcile Re-ID datasets, where each training image corresponds to one ID label, each training image in vehciel search datasets corresponds to multiple ID labels, which is ignored even in person search works that has achieved rapid development.
We propose to leverage the multi-label supervision information to guide image-level identification learning.
Specifically, for each raw frame $\mathbf{x}_t$ in $\mathcal{D}$, we obtain its multi-ID class label $\mathcal{C}_t=\{c_{t}^j\}_{j=1}^{N^{t}}$ through dropping the location annotations of the boxes $b_{t}^{j}$ in $\mathbf{y}_t$.
The label $\mathcal{C}_t$ is then transcribed as the vector $\mathbf{c}_t\in {\{0,1\}}^{C} $, where $\mathbf{c}_t(l)=1$ if the index $l\in \mathcal{C}_t$, else 0.
Given the global image-level feature map $\mathbf{F}_t$ output by the backbone network, we apply a multi-label ID classifier $f_\mathrm{mc}(\cdot)$ to generate the vector of ID class probabilities predicted for $\mathbf{x}_t$, formulated as $\mathbf{p}_t=f_\mathrm{mc}(\mathbf{F}_t) \in {[0,1]}^C$.
The binary cross-entropy (BCE) loss is used as image-level multi-ID classification loss for supervision
\begin{equation}
\begin{aligned}
	  \mathcal{L}_\mathrm{mil}^\mathrm{img} = \sum _t [& -\frac{1}{C}  \sum _{l=1} ^ C \mathbf{c}_t(l)\log(\mathbf{p}_t(l)) \\	
	  & + (1-\mathbf{c}_t(l))\log(1-\mathbf{p}_t(l))].
\end{aligned}
  \label{eq:imagelevel}
\end{equation}
Based on this, the global feature map $\mathbf{F}_t$ can be optimized to focus on the vehicle objects, which can help subsequent vehicle localization and identity distinction.

\textbf{Box-level (instance) ID classification loss.}
Since the proposed vehicle search framework sequentially handles vehicle detection and re-identification, the detection results output by box predictor $f_{\mathrm{box}}(\cdot)$ directly dominate the training efficiency and capability of ID head.
However, due to the fact that \textit{vehicle detection focuses on vehicle commonness and re-identification focuses on vehicle uniqueness}, the detection task may be inconsistent with the re-identification task.
Also, during the early phase of training, when the detection results have a large proportion of errors, ID head learning is more severely affected.
To address these issues, we propose to leverage annotated ground-truth boxes for additional guidance.
Specifically, based on the ground-truth boxes $\mathbf{y}_t$ for $\mathbf{x}_t$, we apply the RoI Align operation and ID head on $\mathbf{F}_t$ to extract ground-truth box-level feature maps $\mathbf{f}_j^{\mathrm{gt}}$. 
For each box feature $\mathbf{f}_{j}^\mathrm{gt}$ of an annotated box $b_{t}^j$ cropped from $\mathbf{F}_t$, it has a corresponding ID label $c_{t}^j$.
We then apply a single-label ID classifier $f_\mathrm{sc}(\cdot)$ to generate the vector of ID class probabilities predicted for $b_{t}^j$, formulated as $\mathbf{p}_{t}^{j}=f_\mathrm{sc}(\mathbf{f}_{j}^\mathrm{gt}) \in {[0,1]}^C$.
The cross-entropy (CE) loss is used as box-level ID classification loss for supervision
\begin{equation}
	  \mathcal{L}_\mathrm{mil}^\mathrm{box} = \sum _t \sum _j -\log(\mathbf{p}_{t}^{j}(c_{t}^{j})).
  \label{eq:boxlevel}
\end{equation}
In total, there are two branches of boxes fed into ID head for training. The branch of ground-truth boxes ensures that ID head can extract discriminative identity features, while the branch of boxes predicted by $f_{\mathrm{box}}(\cdot)$ ensures that ID head can also extract identity features from boxes that do not perfectly surround the vehicles.

\textbf{Feature-level consistency distillation loss.}
In Sec.~\ref{sec:dual}, we introduce CLIP in combination with semantic­-region alignment strategy.
However, CLIP is pre-trained to align whole image features and text features,  and there exists a distribution gap between whole image features and pooled regional features by RoI Align.
Inspired by knowledge distillation~\cite{hinton2015distilling}, we propose to leverage a frozen Re-ID model that has been pre-trained with cropped vehicle patches to alleviate the gap.
Specifically, for the raw frame $\mathbf{x}_t$, we first obtain cropped vehicle patches using ground-truth bounding boxes $\mathbf{y}_t$, and then apply the frozen pre-trained Re-ID model to extract their ID-aware features ${\mathbf{g}}_{j}^\mathrm{gt}$.
For each pair of pre-trained ID feature ${\mathbf{g}}_{j}^\mathrm{gt}$ and pooled regional feature $\mathbf{f}_{j}^\mathrm{gt}$ of ground-truth box $b_{t}^{j}$, we use the $L_1$ loss as feature-level consistency distillation loss to minimize their distance as
\begin{equation}
	  \mathcal{L}_\mathrm{mil}^\mathrm{fea} = \sum _j ||{\mathbf{g}}_{j}^\mathrm{gt} -\mathbf{f}_{j}^\mathrm{gt}||_1.
  \label{eq:fealevel}
\end{equation}
In total, the optimization objective of multi-level identification learning is defined as $\mathcal{L}_\mathrm{mil} = \mathcal{L}_\mathrm{mil}^\mathrm{img} + \mathcal{L}_\mathrm{mil}^\mathrm{box} + \mathcal{L}_\mathrm{mil}^\mathrm{fea}$.

\subsection{Implementation details}
\textbf{Two-stage training.}
In the first pre-training stage, we learn the ID-specific text tokens for vehicle ID classification (in Sec.~\ref{sec:dual}). We keep image encoder and text encoder of CLIP frozen, and optimize the text tokens by contrastive learning loss.
Also, we use a separate Re-ID model for feature-level consistency distillation (in Sec.~\ref{sec:multi}).
With the cropped vehicle patches with ground-truth bounding boxes as input, we pre-train the Re-ID model, \textit{i.e.}, ResNet50~\cite{he2016deep}, on our dataset with standard vehicle Re-ID losses.
In the second training stage, the learned ID-specific text tokens, the pre-trained Re-ID model, and the text encoder of CLIP are frozen, the proposed vehicle search model is trained by $\mathcal{L}=\mathcal{L}_\mathrm{det}+\mathcal{L}_\mathrm{reid}+\mathcal{L}_\mathrm{sra}+\mathcal{L}_\mathrm{mil}$, as shown in Fig.~\ref{fig:overview}.

\textbf{Experimental settings.}
We adopt ResNet50~\cite{he2016deep} in the proposed framework.
The backbone network is composed of layer conv1 to conv4, and both the detection head $f_{\mathrm{det}}(\cdot)$ and ID head $f_{\mathrm{id}}(\cdot)$ are composed of the  remaining layer conv5 without sharing parameters.
The box predictor $f_{\mathrm{box}}(\cdot)$ and box regression $f_{\mathrm{reg}}(\cdot)$ are employed with the corresponding modules in Faster R-CNN~\cite{ren2016faster}.
The multi-label ID classifier $f_\mathrm{mc}(\cdot)$ and single-label ID classifier $f_\mathrm{sc}(\cdot)$ are composed of the pooling operation and a FC layer, have the same structure without sharing parameters.
In the first pre-training stage, CLIP with ResNet50 architecture is adopted, the number of learned text tokens $M$ is set to 4, and the separate pre-trained Re-ID model is initialized with ResNet50 pre-trained on ImageNet.
The $h$, $w$, $d$ of pooled feature maps after RoI Align are 14, 14, 1024.
The $\mathrm{sim}(\cdot,\cdot)$ in Eqs.~\eqref{eq:ali1-1} and \eqref{eq:ali2} is implemented by cosine similarity.
The training batch size is set to 5 and each raw image is resized to $900 \times 1500$ pixels.
All experiments are performed on NVIDIA 3090 GPU with SGD as optimizer.

\section{Vehicle Search Benchmark}
Current works focus on vehicle detection and vehicle re-identification sub-tasks separately, still lacking datasets and evaluation for end-to-end vehicle search.
To promote related researches, we build a new benchmark for this task. 
We collect three large-scale vehicle search datasets, including one real-world dataset CityFlowVS, and two synthetic datasets SynVS-Day and SynVS-All.
The basic statistics of them are shown in Tab.~\ref{tab:dataset}. 
In this section, we will describe the dataset construction in detail, as well as define the evaluation metrics.

\begin{table}
	\centering
	\renewcommand{\arraystretch}{1.15}
	\setlength{\tabcolsep}{1mm}{}
	\caption{Statistics of the proposed vehicle search datasets.}
	\label{tab:dataset}
	\begin{tabular}{@{}lcccc@{}}
		\hline
		\makebox[0.2\linewidth][l]{Datasets} & \makebox[0.175\linewidth][c]{\# Identities} & \makebox[0.175\linewidth][c]{\# Frames} & \makebox[0.175\linewidth][c]{\# Boxes} & \makebox[0.175\linewidth][c]{\# Queries}\\
		\hline
		CityFlowVS & 666 & 18,111 & 50,629 &- \\
		Train & 329 & 6,561 & 17,752 &- \\
		Test & 337 & 11,550 & 32,877 & 1,759 \\
		\hline
		SynVS-Day & 810 & 17,912 & 145,398 & - \\
		Train & 475 & 11,102 & 95,473 & - \\
		Test & 335 & 6,810 & 49,925 & 1,277 \\
		\hline
		SynVS-All & 2,300 & 69,918 & 585,125 & - \\
		Train & 1,457 & 44,273 & 388,087 & - \\
		Test & 843 & 25,645 & 197,038 & 4,038 \\
		\hline
	\end{tabular}
\end{table}

\begin{table*}
	\centering
	\caption{Comparison with the state-of-the-art vehicle Re-ID and person search methods on the proposed vehicle search datasets (\%).} 
	\label{tab:compar}
	\begin{tabular}{@{}llcccccc@{}}
		\hline
		\multirow{2}{*}{\makebox[0.12\textwidth][l]{}} & \multirow{2}{*}{\makebox[0.17\textwidth][l]{Method}} & \multicolumn{2}{c}{CityFlowVS} & \multicolumn{2}{c}{SynVS-Day} & \multicolumn{2}{c}{SynVS-All} \\
		\cline{3-8}
		& & \makebox[0.08\textwidth][c]{mAP} & \makebox[0.08\textwidth][c]{Top-1} & \makebox[0.08\textwidth][c]{mAP} & \makebox[0.08\textwidth][c]{Top-1} & \makebox[0.08\textwidth][c]{mAP} & \makebox[0.08\textwidth][c]{Top-1} \\
		\hline
		\multirow{3}{*}{Vehicle Re-ID} & CLIP-ReID~\cite{li2023clip} & 9.9 & 52.6 & 29.3 & 56.4 & 20.6 & 51.0  \\
		& MSINet~\cite{gu2023msinet} & 6.1 & 26.8 & 18.6 & 35.9 & 8.9 & 17.0 \\
		& MBR~\cite{almeida2023strength} & 10.6 & 75.9 & 16.3 & 80.1 & 10.5 & 64.6  \\
		\hline
		\multirow{3}{*}{Person Search} & SeqNet~\cite{li2021sequential} & 11.7 & 79.8 & 30.2 & 82.6 & 22.5 & 79.8  \\
		& COAT~\cite{yu2022cascade} & 13.3 & 82.5 & 31.6 & 83.0 & 23.7 & 81.3  \\
		& OIMNet++~\cite{lee2022oimnet} & 12.8 & 83.0 & 30.0 & 83.1 & 22.1 & 80.8  \\
		\hline
		& CLIPVehicle & \textbf{14.1} & \textbf{83.8} & \textbf{32.4} & \textbf{84.1} & \textbf{24.6} & \textbf{83.0} \\
		\hline
	\end{tabular}
\end{table*}

\textbf{CityFlowVS.}
We first build a new real-world vehicle search dataset namely CityFlowVS based on a city-scale multi-target multi-camera vehicle tracking dataset CityFlowV2~\cite{tang2019cityflow}, which comprises 3.58 hours (215.03 minutes) of videos collected from 46 cameras spanning 16 intersections in a mid-sized U.S. city.
To build the vehicle search dataset, first, for each raw video, we extracted all frames and filtered out those without vehicles inside.
Then, we sample one in every five frames to obtain 18,111 images in total. The annotation of each image, \textit{i.e.}, the bounding boxes and IDs of the vehicles in it, is inherited from that of the original videos.
After that, we performed statistics for each scenario and divided the dataset according to scenarios, where four (S01, S02, S03, S04) were used for training and one (S05) for testing.
Finally, within the test set, we randomly selected one image of each camera of each vehicle as query, and used the ground-truth bounding boxes to crop out query images.
In total, CityFlowVS consists of 666 vehicles and 50,629 bounding boxes within 18,111 frames.

\textbf{SynVS-Day and SynVS-All.}
We then build two new synthetic vehicle search dataset SynVS-Day and SynVS-All based on a massive synthetic multi-target multi-camera vehicle tracking dataset Synthehicle~\cite{herzog2023synthehicle}, which consists of 17 hours of videos recorded from 340 cameras placed around crossroads and highways in 64 diverse day, rain, dawn, and night scenes.
Following the way that CityFlowVS dataset was constructed, we processed the Synthehicle dataset step by step.
Differently, the ground-truth in Synthehicle contains boxes for both vehicles and pedestrians, and boxes are included also even if they are only visible in one scenario (camera).
Therefore, when constructing our vehicle search datasets, we dropped the annotations for pedestrians and the annotations for vehicles that only appeared in one scenario.
Based on the five town scenarios in the training set of Synthehicle (the annotations of test set of Synthehicle are not publicly available), we used three towns (Town01, Town02, Town03) for training and two towns (Town04, Town05) for testing.
Also, we considered two settings, \textit{i.e.}, 1) including only daytime scenes like most vehicle re-ID datasets and CityFlowVS, named SynVS-Day, 2) including all weather scenes, named SynVS-All, which has more data variety and is more challenging.
Statistically, SynVS-Day consists of 810 vehicles and 145,398 bounding boxes within 17,912 frames.
SynVS-All consists of 2,300 vehicles and 585,125 bounding boxes within 69,918 frames.
More detailed statistics about the training/test splitting are shown in Tab.~\ref{tab:dataset}.

\textbf{Evaluation metrics.}
Following the evaluation protocol in vehicle Re-ID~\cite{zhang2024towards} and person search~\cite{zhang2024towards}, we adopt the mean Average Precision (mAP) and Cumulative Matching Characteristics (CMC Top-1) as evaluation metrics to measure the vehicle search performance.
Given a query vehicle image and gallery frames, we calculate the similarity score between the query and all detected boxes in gallery frames to obtain a ranking list.
The gallery box would be considered as a true positive if its IoU with the ground-truth bounding box that shares the same ID label with the query is larger than 0.5.
This way, the evaluation of both detection and re-identification results are taken into consideration.

\section{Experimental Results}
\subsection{Setup}

\begin{table*}
	\centering
	\caption{Comparison with state-of-the-art vehicle Re-ID and person search methods under different scenes on SynVS-All dataset (\%).}
	\label{tab:scene}
	\begin{tabular}{@{}llcccccccc@{}}
		\hline
		\multirow{2}{*}{} & \multirow{2}{*}{Method} & \multicolumn{2}{c}{Day} & \multicolumn{2}{c}{Dawn} & \multicolumn{2}{c}{Rain} & \multicolumn{2}{c}{Night} \\
		\cline{3-10}
		& & \makebox[0.06\textwidth][c]{mAP} & \makebox[0.06\textwidth][c]{Top-1} & \makebox[0.06\textwidth][c]{mAP} & \makebox[0.06\textwidth][c]{Top-1} & \makebox[0.06\textwidth][c]{mAP} & \makebox[0.06\textwidth][c]{Top-1} & \makebox[0.06\textwidth][c]{mAP} & \makebox[0.06\textwidth][c]{Top-1} \\
		\hline
		\multirow{3}{*}{Vehicle Re-ID} & CLIP-ReID~\cite{li2023clip} & 31.3 & 50.6 & 34.1 & 54.6 & 27.8 & 55.9 & 18.3 & 46.2  \\
		& MSINet~\cite{gu2023msinet} & 14.0 & 21.0 & 14.7 & 24.0 & 14.3 & 24.9 & 8.5 & 14.7 \\
		& MBR~\cite{almeida2023strength} & 13.7 & 69.6 & 11.8 & 70.2 & 15.9 & 67.4 & 9.7 & 51.8 \\
		\hline
		\multirow{3}{*}{Person Search} & SeqNet~\cite{li2021sequential} & 30.0 & 82.4 & 32.7 & 81.4 & 28.6 & 82.7 & 18.8 & 73.3 \\
		& COAT~\cite{yu2022cascade} & 31.5 & 83.5 & 34.1 & 83.0 & 30.4 & 83.6 & 20.0 & 75.3 \\
		& OIMNet++~\cite{lee2022oimnet} & 30.8 & 82.7 & 33.5 & 81.9 & 28.4 & 83.1 & 18.6 & 74.7 \\
		\hline
		& CLIPVehicle & \textbf{32.6} & \textbf{85.0} & \textbf{36.2} & \textbf{84.2} & \textbf{32.1} & \textbf{85.5} & \textbf{21.6} & \textbf{77.3} \\
		\hline
	\end{tabular}
\end{table*}

We evaluate the proposed CLIPVehicle and the state-of-the-art methods on the CityFlowVS, SynVS-Day and SynVS-All datasets, respectively.
Note that, as a newly proposed problem, there is no existing method directly designed for end-to-end vehicle search.
In order to compare with previous methods as much as possible, we include two categories of methods in the experiments. The first category is vehicle Re-ID methods, including  CLIP-ReID~\cite{li2023clip}, MSINet~\cite{gu2023msinet}, and MBR~\cite{almeida2023strength}.
Note that, the vehicle Re-ID methods are designed with vehicle patches as input, we use the cropped vehicles with ground-truth boxes during training and with detection results predicted by our method during testing.
The second category of comparison methods are used for person search, \textit{i.e.}, SeqNet~\cite{li2021sequential}, COAT~\cite{yu2022cascade}, and OIMNet++~\cite{lee2022oimnet}.

\subsection{Comparison results}

\textbf{Main result analysis.}
Table~\ref{tab:compar} shows the comparison results between the proposed method and state-of-the-art vehicle Re-ID and person search methods.
We can see that the vehicle Re-ID methods all achieve relatively poor performance, which may be attributed to the fact that they do not consider the inaccurate vehicle detection results, and just learn to extract identity features from well-annotated vehicle patches. 
As a result, Re-ID performance of such methods has been affected when facing imperfect detected boxes.
This indicates that, the vehicle Re-ID methods, \textit{not only require the prior algorithm and storage for the detection, but also provide poor results}, for vehicle search problem.
The methods for person search, a task closely related to this work, consider both detection and Re-ID, and consequently achieve better results.
The proposed end-to-end unified vehicle search framework CLIPVehicle achieves the highest mAP and Top-1 accuracy on both real-world (CityFlowVS) and synthetic (SynVS-Day and SynVS-All) datasets.
Specifically, we outperform the vanilla baseline SeqNet (using the same backbone as ours) over two percentage points for mAP score on all three datasets.
Also, CLIPVehicle outperforms the state-of-the-art approaches COAT and OIMNet++, both of which are based on SeqNet with new modules.
Finally, we can see that the overall performance of the proposed vehicle search, especially the mAP score, is not good enough. This demonstrates that \textit{this problem is quite challenging and leaves a lot of space for further research}.

\textbf{Attribute-aware result analysis.}
As the SynVS-All dataset consists of different weather scenes, \textit{i.e.}, day, rain, dawn, and night, for further in-depth analysis, we report the results of comparison methods and ours under specific weather settings in Tab.~\ref{tab:scene}.
As observed, all methods work better in the day and dawn scenes and perform slightly worse in the rain and night scenes.
Especially in night, performance of all methods drops significantly, which inspires us that how to achieve vehicle search in low-light scenes is one of the important research tasks in the future.


\begin{table}
	\centering
	\caption{Ablation study of each component on CityFlowVS (\%).}
	\label{tab:ablation}
	\setlength{\tabcolsep}{1.75mm}{}
	\begin{tabular}{@{}lcc|ccc|cc@{}}
		\hline
		&	\multicolumn{2}{c|}{Dual-granularity} & \multicolumn{3}{c|}{Multi-level} & \multirow{2}{*}{mAP} & \multirow{2}{*}{Top-1} \\
		\cline{2-6}
		&	\makebox[0.06\textwidth][c]{Obj} & \makebox[0.06\textwidth][c]{ID} & \makebox[0.04\textwidth][c]{Image} & \makebox[0.04\textwidth][c]{Box} & \makebox[0.04\textwidth][c]{Feature} &  &  \\
		\hline
		\ding{192} &	 & & & & & 11.7 & 79.8  \\
		\ding{193}&	\checkmark & & & & & 12.2 & 81.4  \\
		\ding{194}&	& \checkmark & & & & 12.7 & 81.5  \\
		\ding{195}&	\checkmark & \checkmark & & & & 13.0 & 82.3 \\
		\ding{196}&	& & \checkmark & & & 11.9 & 80.6 \\
		\ding{197}&	& & & \checkmark & & 12.4 & 82.2 \\
		\ding{198}&	& & & & \checkmark & 12.4 & 81.4 \\
		\ding{199}&	& & \checkmark & \checkmark & \checkmark & 12.8 & 82.5  \\
		\ding{200}&	\checkmark & \checkmark & \checkmark & \checkmark & \checkmark & \textbf{14.1} & \textbf{83.8}  \\
		\hline
	\end{tabular}
\end{table}

\subsection{Ablation study}

Table~\ref{tab:ablation} reports the ablation studies of each component of the proposed CLIPVehicle on CityFlowVS.

\textbf{Effectiveness of the dual-granularity semantic-region alignment} (Sec.~\ref{sec:dual}).
We can first see from comparing the results between line \ding{192} and line \ding{195} that, with the introduction of text prompt-based semantic-region alignment, the mAP and Top-1 are improved by 1.3\% and 2.5\%, respectively.
Furthermore, when applied alone, both object-granularity (line \ding{193}) and ID-granularity (line \ding{194}) text prompts improve vehicle search performance, which are verified to be effective for guiding foreground-background discrimination in detection sub-task, and identity classification in Re-ID sub-task, respectively.
We also compare different ID-granularity text prompts, \textit{i.e.}, with and without attribute-based descriptions, in Tab.~\ref{tab:IDtext}, where the results confirm the effectiveness of vehicle attribute cues for identity classification.

\textbf{Effectiveness of the multi-level identification learning} (Sec.~\ref{sec:multi}).
As seen, with the coarse-to-fine learning strategy applied, the mAP and Top-1 are improved by 1.1\% and 2.7\%, respectively (comparing the results between line \ding{192} and line \ding{199}).
More specifically, from line \ding{196} to line \ding{198} in Tab.~\ref{tab:ablation} report the results of using image-level (global) multi-ID classification loss, box-level (instance) ID classification loss, and feature-level consistency distillation loss individually, which are all verified to enhance the vehicle search performance.

\begin{table}
	\centering
	\caption{Comparison of different ID-granularity text prompts (\%).}
	\label{tab:IDtext}
	\begin{tabular}{@{}lcc@{}}
		\hline
		\makebox[0.33\textwidth][l]{ID-granularity text prompt type} & mAP & Top-1 \\
		\hline
		`A photo of a $X$ vehicle.' & 12.1 & 80.6  \\
		\hline
		\makecell[l]{`A photo of a $X$ vehicle with \\ $X$ color and $X$ type.'} & 12.7 & 81.5 \\
		\hline
	\end{tabular}
\end{table}

\begin{table} [t]
	\centering
	\caption{Combination of the proposed strategy with state-of-the-art person search models (\%).}
	\label{tab:comb}
	\begin{tabular}{@{}lccc@{}}
		\hline
		\makebox[0.2\textwidth][l]{Methods} & \makebox[0.06\textwidth][c]{mAP} & \makebox[0.06\textwidth][c]{Top-1} & \makebox[0.06\textwidth][c]{FPS} \\
		\hline
		CLIPVehicle (w SeqNet~\cite{li2021sequential}) & 14.1 & 83.8 & 10.1 \\
		CLIPVehicle (w COAT~\cite{yu2022cascade}) & 15.2 & 85.8 & 6.6 \\
		CLIPVehicle (w OIMNet++~\cite{lee2022oimnet}) & 14.3 & 85.6 & 7.3 \\
		\hline
	\end{tabular}
\end{table}

\subsection{Model adaptability}

Finally, we explore the combination of the proposed models with state-of-the-art approaches.
Specifically, in above experiments, we select SeqNet~\cite{li2021sequential} as baseline since it is a vanilla architecture with faster inference speed, which is important since vehicle search needs to retrieve specific target from massive galleries.
To investigate the adaptation of the proposed method, we further equip it into other approaches COAT and OIMNet++.
As seen in Tab.~\ref{tab:comb}, with our model added, all baseline models achieve higher performance.
We compare the inference speed in frames per second (FPS) in the last column.
Note that, the proposed strategy is applied only during training phase, that is, \textit{our method does not affect the inference speed of baseline models}.
In conclusion, the proposed method is effective, and flexible enough to equip with different object search baselines for higher performance and efficiency.

\section{Conclusion}
In this work, we focus on end-to-end vehicle search in a unified framework, which integrates the vehicle detection and re-identification. Although more practical, it is more challenging.
To handle this problem, we have designed a novel method CLIPVehicle including a dual-granularity semantic-region alignment module and a multi-level vehicle identification learning strategy, to jointly and effectively localize and discriminate the vehicles in the search gallery.
We have also constructed a new benchmark for training and evaluation of vehicle search, extensive experimental results on which demonstrate that our method achieves the state-of-the-art performance.
Through the above efforts, we aim to pave the way for the research of this new and practical problem, which has the potential to be applied in the real-world intelligent transportation systems.

\bibliographystyle{IEEEtran}
\bibliography{IEEEabrv,manuscript}

%
%

\vfill

\end{document}